\documentclass[manuscript]{acmart}
\AtBeginDocument{%
  }

\setcopyright{acmlicensed}
\copyrightyear{2026}
\acmYear{2026}
\acmDOI{XXXXXXX.XXXXXXX}
\acmConference[Conference acronym 'XX]{Make sure to enter the correct
  conference title from your rights confirmation email}{June 03--05,
  2026}{Woodstock, NY}
\acmISBN{978-1-4503-XXXX-X/2018/06}

\usepackage{multirow} 

\begin{document}

\title{Benchmarking Vision-Language-Action Models on SO-101: Failure and Recovery Analysis}
%Benchmarking Vision-Language-Action Models on SO-101: Failure and Recovery Analysis
\author{Yi Yu}
\affiliation{%
  \institution{Graduate School of Advanced Science and Engineering, Hiroshima University}
  \city{Higashi-Hiroshima}
  \country{Japan}
}
\email{yiyu@hiroshima-u.ac.jp}

\author{Xinchuan Qiu}
\affiliation{%
  \institution{Graduate School of Advanced Science and Engineering, Hiroshima University}
  \city{Higashi-Hiroshima}
  \country{Japan}
}
\email{qiuxinchuan2025@163.com}

\renewcommand{\shortauthors}{Yu et al.}

\begin{abstract}
Vision-Language-Action (VLA) models have demonstrated strong generalization in robotic manipulation, yet existing evaluations are primarily conducted in simulation or on expensive robotic platforms, leaving their robustness on affordable real-world robots largely unexplored. We present a standardized real-world benchmark for evaluating representative VLA and imitation learning policies on the low-cost SO-101 robotic platform. The benchmark comprises four representative manipulation tasks together with unified evaluation protocols, enabling systematic comparison under embodiment uncertainty. Using real-world teleoperated demonstrations, we fine-tune and evaluate $\pi_{0.5}$, SmolVLA, Wall-X, and ACT directly on the physical platform. Beyond conventional task success rates, the benchmark incorporates a structured failure taxonomy, semantic- and execution-level failure decomposition, and recovery-aware evaluation metrics to characterize policy robustness. Experimental results show that stronger pretrained VLA policies generally outperform the imitation learning baseline, although performance remains highly task-dependent under low-cost robotic deployment conditions. 
Execution instability emerges as the dominant failure source, while recovery capability varies substantially across architectures. These results highlight the importance of failure and recovery analysis beyond binary task success and establish SO-101 as a practical benchmark for evaluating embodied AI systems under realistic low-cost robotic deployment conditions.

\end{abstract}


%\ccsdesc[Computer systems organization → Robotics}


%\received{20 February 2026}
%\received[revised]{12 March 2026}
%\received[accepted]{5 June 2026}

\maketitle

\section{Introduction}

Recent advances in Vision-Language-Action (VLA) models~\cite{RT2, palm_e, OpenVLA, Octo} have significantly improved the ability of robotic systems to generalize across diverse manipulation tasks specified through natural language instructions. Leveraging large-scale robot datasets such as Open X-Embodiment~\cite{OpenXEmbodiment}, pretrained policies have demonstrated promising performance across a wide range of embodied manipulation settings. 
However, most existing evaluations are conducted either in simulation environments or on expensive research-grade robotic platforms~\cite{RLBench, RoboSuite}, where controlled dynamics and high-precision hardware often obscure challenges that arise during real-world deployment.

In contrast, deployment on low-cost robotic platforms introduces substantially noisier embodiment conditions, including actuation inaccuracies, unstable grasping, sensing limitations, and accumulated execution drift. These factors can significantly affect policy behavior and expose failure patterns that are difficult to observe under controlled evaluation settings. Despite the growing accessibility of low-cost robotic systems, it remains unclear to what extent current VLA policies can maintain robust performance under realistic embodiment perturbations. Consequently, there is a need for evaluation frameworks that move beyond simulation-based assessments and systematically characterize policy robustness under real-world deployment conditions.

Existing benchmark suites such as RLBench~\cite{RLBench} and RoboSuite~\cite{RoboSuite} have significantly advanced robotic policy evaluation by providing standardized simulation environments and reproducible evaluation protocols. However, these platforms are primarily simulation-based and therefore cannot fully capture embodiment uncertainty encountered during real-world deployment. Meanwhile, recent real-world evaluations have largely focused on high-end research platforms, leaving the robustness of VLA policies under low-cost deployment conditions relatively unexplored. Consequently, there remains a need for practical real-world benchmarks that systematically evaluate policy robustness under realistic embodiment noise and hardware constraints.

In this work, we argue that low-cost robotic platforms should not be viewed merely as constrained deployment environments, but as valuable stress-test settings for embodied AI systems. Under embodiment uncertainty, such platforms reveal characteristic failure behaviors, including grasp instability, repetition loops, state mismatch, and precision misalignment, which are often underrepresented in conventional benchmarks. Understanding these failure patterns is essential for evaluating the practical robustness of modern VLA policies beyond aggregate task success rates.

To address this gap, we introduce a standardized real-world benchmark built on the low-cost SO-101 platform~\cite{so101}. The benchmark consists of four representative manipulation tasks together with unified evaluation protocols, enabling systematic comparison of policy behavior under identical deployment conditions. Using real-world teleoperated demonstrations, we fine-tune and evaluate representative VLA policies, including $\pi_{0.5}$~\cite{pi05}, SmolVLA~\cite{SmolVLA}, and Wall-X~\cite{WallX}, as well as the imitation learning baseline ACT~\cite{ACT}. Beyond conventional task success rates, the benchmark incorporates a structured failure taxonomy, semantic- and execution-level failure decomposition, and recovery-aware evaluation metrics to provide a more comprehensive characterization of policy robustness.

Our experiments show that stronger pretrained VLA policies generally outperform the imitation learning baseline under low-cost deployment conditions. However, performance remains highly task-dependent, and execution instability emerges as the dominant source of failure across all evaluated methods. Furthermore, recovery capability varies substantially across architectures, indicating that task completion performance alone is insufficient to fully characterize embodied robustness. These findings suggest that failure dynamics and recovery behavior provide important complementary signals beyond binary success metrics when evaluating embodied manipulation policies. The main contributions of this work are summarized as follows:

\begin{itemize}

\item We introduce a standardized real-world benchmark for evaluating Vision-Language-Action (VLA) policies on the low-cost SO-101 robotic platform, providing an accessible and reproducible testbed for embodied AI research under realistic deployment conditions.

\item We propose a robustness-oriented evaluation framework that combines task success rates, a structured failure taxonomy, semantic- and execution-level failure analysis, and recovery-aware evaluation metrics, enabling a more comprehensive assessment of policy robustness beyond binary task completion.

\item We conduct a systematic real-world evaluation of $\pi_{0.5}$, SmolVLA, Wall-X, and ACT under identical deployment conditions, providing a direct comparison of representative VLA and imitation learning policies on a low-cost robotic platform.

\item We provide empirical insights into the robustness limitations of current VLA policies under low-cost real-world deployment, highlighting execution instability as a dominant failure source and demonstrating the importance of recovery-aware evaluation beyond task success rates.

\end{itemize}

\section{Methodology}

This section presents the evaluation methodology used to benchmark Vision-Language-Action (VLA) and imitation learning policies on a low-cost real-world robotic platform. The proposed pipeline consists of execution-centric task design, teleoperated data collection, task-specific policy adaptation, and standardized real-world evaluation. The benchmark is designed to assess not only task completion performance but also failure behavior, execution stability, and recovery capability under embodiment uncertainty.

\subsection{Robotic Platform}

All data collection, task-specific policy adaptation, and real-world evaluations are conducted on the SO-101 robotic platform, an open-source low-cost tabletop manipulator designed for embodied AI research. Teleoperated demonstrations collected on the platform are used for task-specific fine-tuning, and all policies are subsequently evaluated under identical deployment conditions. Compared with research-grade robotic systems such as Franka Panda~\cite{franka_panda} and WidowX~\cite{widowx}, the SO-101 operates under substantially stronger embodiment constraints, including limited actuator precision, reduced control stability, lower payload capacity, and noisier motion execution. These characteristics make it a suitable platform for studying the gap between benchmark performance and real-world deployment robustness.

The platform introduces stochasticity into the perception-action loop through actuator noise, joint backlash, control latency, trajectory jitter, limited positional repeatability, and imperfect calibration between visual observations and executed actions. As a result, small execution errors can accumulate over time, particularly in long-horizon and contact-rich manipulation tasks. Such hardware-induced perturbations substantially influence policy behavior and amplify embodiment-specific effects on task outcomes. Consequently, task success alone may not fully capture policy robustness under realistic deployment conditions.
This motivates an execution-centric evaluation perspective that focuses on directly observable execution properties, including control fidelity, grounding robustness, temporal consistency, and robustness under embodiment noise. In addition to task completion performance, recovery capability is also evaluated as a complementary indicator of policy robustness under execution disturbances.

Despite its low-cost design, the SO-101 supports a range of tabletop manipulation behaviors, including grasping, object transport, object sorting, packing, and insertion. These capabilities enable systematic evaluation of embodied policies across tasks requiring both stable physical control and robust visuomotor coordination.
Overall, the SO-101 serves as a practical and accessible testbed for studying policy robustness, failure dynamics, and recovery behavior under realistic low-cost deployment conditions. Its physical limitations provide a useful lens for exposing failure modes that are often hidden in simulation-based evaluations.

\subsection{Task Design Rationale}

\subsubsection{From Semantic Complexity to Execution-Centric Evaluation}

A common approach to benchmark design is to organize tasks according to increasing semantic complexity, where instructions become progressively more compositional and cognitively demanding. However, preliminary experiments revealed a potential limitation of this formulation: successful task completion does not necessarily indicate that the intended semantic capability has been acquired. For example, tasks involving object attributes or spatial relations may sometimes be solved through learned visuomotor correlations that are sufficient for successful execution under specific environmental conditions. Consequently, performance differences across task levels may not reliably reflect differences in the underlying semantic competencies assumed by the benchmark design.

Motivated by this observation, we adopt an execution-centric evaluation framework. Rather than organizing tasks according to assumed semantic difficulty, tasks are designed to probe directly observable execution properties that can be consistently measured during real-world deployment. As summarized in Table~\ref{tab:task}, the benchmark consists of four representative manipulation tasks spanning distinct execution dimensions, including control fidelity, grounding robustness, temporal consistency, and robustness under embodiment noise. This formulation enables systematic analysis of execution stability, failure behavior, and recovery capability under realistic low-cost deployment conditions.

\begin{table}[t]
\centering
\caption{Execution-centric benchmark tasks.}
\label{tab:task}
\begin{tabular}{lll}
\toprule
Task & Instruction & Evaluated Capability \\
\midrule
Pen Transfer &
Move the pen from left to right. &
Control fidelity \\

Selective Color Sorting &
Put the pink block in the plate and remove the others. &
Grounding robustness \\

Multi-Object Packing &
Place all snacks into the cardboard box. &
Temporal consistency \\

Precision Pen Placement &
Pick up the pen and insert it into the pen holder. &
Embodiment-noise robustness \\
\bottomrule
\end{tabular}
\end{table}

\subsubsection{Execution-Centric Tasks}
\label{sec:Execution-CentricTasks}

The limitations of semantic-level evaluation become particularly pronounced on low-cost robotic platforms, where embodiment noise and execution variability can substantially affect task outcomes. To address this challenge, we construct an execution-centric benchmark consisting of four representative manipulation tasks designed to probe complementary aspects of policy execution.
Specifically, \textit{Pen Transfer} evaluates low-level control fidelity and grasp execution; \textit{Selective Color Sorting} assesses language-conditioned object grounding under distractor interference; \textit{Multi-Object Packing} evaluates long-horizon execution stability and error accumulation; and \textit{Precision Pen Placement} measures robustness under fine-grained positional and contact constraints.

The tasks are selected to balance experimental reproducibility, demonstration availability, and coverage of distinct execution challenges. Together, they span a range of manipulation behaviors requiring stable control, accurate grounding, long-horizon planning, and precise physical interaction. Notably, \textit{Precision Pen Placement} serves as a stricter variant of \textit{Pen Transfer}, requiring accurate alignment and insertion under tighter positional constraints. This imposes substantially greater demands on control precision and execution stability. Overall, the benchmark provides a structured evaluation suite for analyzing policy behavior under realistic low-cost deployment conditions, with particular emphasis on robustness, failure modes, error accumulation, and recovery behavior.

\subsection{Data Collection}

All demonstrations are collected directly on the physical SO-101 platform through human teleoperation using end-effector control. During data collection, operators execute task trajectories while the system records synchronized observation-action pairs, including RGB observations, language instructions, and low-level robot control commands.
By collecting demonstrations entirely on the target hardware platform, both policy adaptation and evaluation are conducted under consistent embodiment conditions. This design eliminates reliance on simulation-generated data and minimizes discrepancies between training and deployment environments.

For each task, 100 teleoperated demonstrations are collected. To improve generalization while preserving comparability, object poses and scene layouts are randomized within predefined workspace constraints. This controlled variability exposes policies to diverse execution conditions without altering task semantics.
Following adaptation, all policies are deployed directly on the physical SO-101 platform and evaluated under identical real-world conditions. Each model-task pair is evaluated over 20 independent hardware trials to assess task success, failure behavior, and recovery capability. In total, 400 teleoperated demonstrations are collected for task-specific fine-tuning. For evaluation, we conduct 320 real-world episodes across four models and four manipulation tasks under identical conditions. Specifically, each model-task pair is evaluated over 20 independent episodes, resulting in a total of 320 evaluation episodes. All collected demonstration datasets have been publicly released to support reproducible research~\cite{SO101_Dataset}.

\subsection{Evaluated Policies and Task Adaptation}

\subsubsection{Evaluated Policies}

We evaluate four representative manipulation policies spanning both VLA models and classical imitation learning approaches: $\pi_{0.5}$, SmolVLA, Wall-X, and ACT.
Among these methods, $\pi_{0.5}$, SmolVLA, and Wall-X are recent VLA policies that generate actions conditioned on visual observations and language instructions. In contrast, ACT is an action-chunking transformer trained via behavior cloning and serves as a strong imitation learning baseline without vision-language pretraining. Together, these models represent diverse architectural designs, scales, and prior knowledge sources, enabling systematic comparison under identical low-cost deployment conditions.

\subsubsection{Policy Adaptation}

For each benchmark task, all policies are adapted using the task-specific demonstration datasets described in Section~\ref{sec:Execution-CentricTasks}. All methods are trained under consistent data splits, task definitions, and evaluation protocols. Each policy is fine-tuned following its original training paradigm and implementation details, including flow-matching objectives for diffusion-based policies and reconstruction- or token-based objectives for imitation learning baselines.

To ensure fair comparison across heterogeneous architectures, only minimal interface-level adaptations are introduced to align each policy with the SO-101 observation format and action space. No modifications are made to model architectures, optimization procedures, or learning objectives. This design preserves the original inductive biases and learning characteristics of each method while enabling deployment on a shared robotic embodiment. Consequently, observed performance differences can be largely attributed to variations in model architecture and learning paradigm rather than differences in data, hardware, or training configurations.

\subsection{Evaluation Metrics}

We evaluate all policies using complementary metrics that characterize task success, failure behavior, semantic- and execution-level failure characteristics, and recovery capability. All metrics are computed from multiple independent trials conducted on the physical platform under identical experimental conditions. Task definitions and success/failure criteria for the proposed benchmark in Table~\ref{tab:task_definition}.

\subsubsection{Success Rate}

Success rate measures the proportion of trials in which the policy completes the target task successfully:

\[
\text{Success Rate}
=
\frac{N_{\text{success}}}
{N_{\text{total}}}.
\]

A trial is considered successful only when the final object state satisfies the predefined task completion criteria without any human intervention or manual correction.

\subsubsection{Failure Taxonomy}
\label{sec:FailureMode}

To provide a more detailed characterization of policy behavior beyond binary success metrics, failed executions are categorized into four representative failure modes:

\begin{itemize}

\item \textbf{Grasp Instability}: Failure to establish or maintain a stable grasp, resulting in unsuccessful object acquisition or unintended object slippage during manipulation.

\item \textbf{Repetition Loop}: Repeated, oscillatory, or redundant action sequences that fail to make meaningful progress toward task completion.

\item \textbf{State Mismatch}: Failure caused by an inconsistency between the inferred task state of the robot and the actual environment state. The policy continues executing subsequent actions based on an incorrect internal assumption without detecting or correcting the discrepancy (e.g., attempting object transport after a failed grasp).

\item \textbf{Precision Misalignment}: Failure to achieve the positioning, alignment, or insertion accuracy required under strict spatial constraints despite otherwise correct task execution.

\end{itemize}

Each failed trial is assigned a single primary failure mode corresponding to the dominant cause of failure. The distribution of failure modes is reported for each model and task to facilitate qualitative comparison of policy behavior.

\subsubsection{Semantic- and Execution-Level Failure Decomposition}

To provide a higher-level interpretation of policy failures, the fine-grained failure taxonomy is further aggregated into semantic-related and execution-related categories. Semantic-related failures are represented by \textit{State Mismatch}, which reflects incorrect scene understanding, task grounding errors, or inconsistencies between the internal task representation of the policy and the observed environment state.

Execution-related failures are computed as the average occurrence of \textit{Grasp Instability} and \textit{Repetition Loop}, which primarily capture low-level execution instability during manipulation, including unstable grasping, motion control errors, and failures to make meaningful progress toward task completion.

\textit{Precision Misalignment} is excluded from this aggregation because it mainly reflects fine-grained spatial alignment errors that arise in a subset of task-specific scenarios, rather than a general execution instability pattern shared across all benchmark tasks. For each evaluated policy, semantic-level and execution-level failure rates are computed by aggregating the corresponding failure categories over all failed evaluation episodes. This coarse-grained decomposition provides complementary insights beyond individual failure categories by distinguishing failures primarily associated with high-level semantic reasoning from those caused by low-level execution deficiencies.

\subsubsection{Recovery Rate}

Recovery rate measures the ability of a policy to recover from intermediate execution failures and subsequently continue task execution:

\[
\text{Recovery Rate}
=
\frac{N_{\text{successful recovery}}}
{N_{\text{recovery opportunity}}}.
\]

Here $N_{\text{successful recovery}}$ denotes the number of failure events from which a policy successfully restores a valid task state and resumes progress toward task completion, while $N_{\text{recovery opportunity}}$ denotes the total number of failure events encountered during evaluation.

A recovery is considered successful when a policy restores a valid task state after an execution failure and continues task execution without external intervention. This metric complements success rate by evaluating not only whether a task is ultimately completed, but also the policy ability to recover from execution failures caused by embodiment noise, perception uncertainty, and environmental disturbances commonly observed in low-cost robotic platforms.

\section{Experimental Results}

\begin{table}[t]
\centering
\caption{Task definitions and success/failure criteria for the proposed benchmark}
\label{tab:task_definition}
\begin{tabular}{p{2.8cm}p{5.2cm}p{5.0cm}}
\toprule
Task & Success Case & Failure Case \\
\midrule

\textbf{Pen Transfer}
&
Successfully grasps the object and transfers it to the target location
&
Grasp failure or incorrect placement
\\

\textbf{Selective Color Sorting}
&
Correctly identifies the target pink object and removes distractors
&
Incorrect object selection or manipulation errors
\\

\textbf{Multi-object Packing}
&
Successfully packs all objects into the target box
&
Missing objects or incomplete packing
\\

\textbf{Precision Pen Placement}
&
Successfully inserts the object into the target holder
&
Insertion misalignment or repeated unsuccessful attempts
\\

\bottomrule
\end{tabular}
\end{table}

We evaluate all policies on the four execution-centric manipulation tasks described in Section~\ref{sec:Execution-CentricTasks} : Pen Transfer, Selective Color Sorting, Multi-object Packing, and Precision Pen Placement. All experiments are conducted through real-world deployment on the SO-101 platform under identical hardware configurations and evaluation protocols to ensure a fair comparison.
Each policy is fine-tuned using 100 human demonstration trajectories per task.
Evaluation is performed at the model-task pair level. For each model-task pair, we conduct 20 independent real-world evaluation episodes, resulting in a total of 320 evaluation episodes across four models and four benchmark tasks.

We first report overall task success rates, followed by a detailed analysis of failure modes using the proposed taxonomy and semantic-level and execution-level failure analysis. We then evaluate policy robustness through recovery rates under execution disturbances, episode-level task performance, and representative trajectories that visualize both successful and failed executions.

\subsection{Success Rate}

Table~\ref{tab:success_rate} summarizes task success rates across all evaluated models and tasks. Overall, VLA-based policies generally demonstrate comparable or superior performance to the ACT baseline across most tasks, particularly in tasks requiring language grounding and long-horizon sequential execution. Among all evaluated methods, $\pi_{0.5}$ achieves the highest overall success rate.
Although SmolVLA achieves a slightly lower average success rate than ACT (32.5\% vs.\ 33.75\%), the difference corresponds to only a small number of successful trials and may be influenced by the limited evaluation budget (20 episodes per task). These results suggest that policy performance is highly task-dependent under low-cost robotic deployment conditions characterized by embodiment noise and execution uncertainty.

From a task-level perspective, \textit{Selective Color Sorting} is consistently the most challenging benchmark, with all evaluated policies achieving very low success rates (0--10\%). In contrast, \textit{Pen Transfer} achieves consistently high success rates (70--95\%) across all evaluated policies, indicating that simple pick-and-place manipulation is comparatively robust under the current benchmark setting. Meanwhile, \textit{Multi-object Packing} exhibits the largest performance variation across policies, suggesting that tasks involving long-horizon sequential manipulation provide stronger discriminative power for evaluating current VLA policies under real-world deployment.

\begin{table}[t]
\centering
\caption{Task Success Rate (\%) across all evaluated tasks.}
\label{tab:success_rate}
\begin{tabular}{lccccc}
\toprule
Model &
Pen Transfer &
Color Sorting &
Object Packing &
Pen Placement &
Average \\
\midrule
 ACT & 75\% & 0\% & 10\% & 50\% & 33.75\% \\
SmolVLA     & 70\% & 5\% & 10\% & 45\% & 32.5\% \\
Wall-X      & 95\% & 0\% & 30\% & 80\% & 51.25\% \\
$\pi_{0.5}$ & 95\% & 10\% & 55\% & 65\% & 56.25\% \\
\bottomrule
\end{tabular}
\end{table}

\subsection{Failure Mode Analysis}

To better understand policy behavior beyond binary success rates, we analyze the distribution of failure modes using the proposed failure taxonomy. Table~\ref{tab:failure_modes} reports the percentage distribution of dominant failure categories over all failed trials. The results reveal distinct and systematic failure signatures across models, rather than uniform or random execution errors. ACT exhibits extremely high rates of \textit{grasp instability}, \textit{repetition loops}, and \textit{state mismatch}, indicating failures across both perception and action consistency. In contrast, VLA-based models tend to reduce \textit{state mismatch} but still suffer from persistent \textit{grasp instability} and \textit{repetition loops}. Notably, $\pi_{0.5}$ achieves the lowest \textit{state mismatch} rate among all methods, suggesting improved global state consistency. Meanwhile, Wall-X eliminates \textit{precision misalignment} entirely but still exhibits high levels of low-level execution failures.

The presence of 0\% entries should be interpreted in the context of finite evaluation samples. Failure mode distributions are computed over failed episodes only and reflect the empirical distribution of dominant failure types observed within each model-task setting. Consequently, certain failure categories may not appear under the current evaluation budget, particularly when failure trajectories are dominated by a small subset of recurring behaviors such as grasp instability or repetition loops.

\begin{table}[t]
\centering
\caption{Failure mode distribution (\%) over failed trials.}
\label{tab:failure_modes}
\begin{tabular}{lcccc}
\toprule
Model &
Grasp Instability &
Repetition Loop &
State Mismatch &
Precision Misalignment \\
\midrule
 ACT & 94.34\% & 92.45\% & 98.11\% & 15.09\% \\
SmolVLA     & 92.59\% & 94.44\% & 70.37\% & 7.41\% \\
Wall-X      & 100\% & 100\% & 61.54\% & 0\% \\
$\pi_{0.5}$ & 91.43\% & 91.43\% & 45.71\% & 14.29\% \\
\bottomrule
\end{tabular}
\end{table}

\subsection{Semantic-Level and Execution-Level Failure Analysis}
\label{sec:failure_analysis}

To provide a higher-level interpretation of policy failures, we aggregate fine-grained failure modes into two categories: semantic-related failures and execution-related failures. Semantic-related failures are primarily represented by \textit{state mismatch}, which reflects incorrect scene understanding, task grounding errors, or inconsistencies between the internal state representation of the policy and the observed environment. Execution-related failures are defined as the average occurrence of \textit{grasp instability} and \textit{repetition loop}, which reflect low-level execution instability during manipulation, including motion execution errors and failures to make progress toward task completion. We exclude \textit{precision misalignment} from this aggregation because it primarily corresponds to task-specific spatial alignment errors rather than a general execution instability pattern shared across benchmark tasks. This grouping enables a coarse-grained analysis of whether failures are primarily driven by high-level semantic reasoning errors or low-level execution deficiencies, providing complementary insights beyond individual failure categories.

\begin{figure}[t]
    \centering
    \includegraphics[width=0.6\linewidth,
    height=4.4cm]{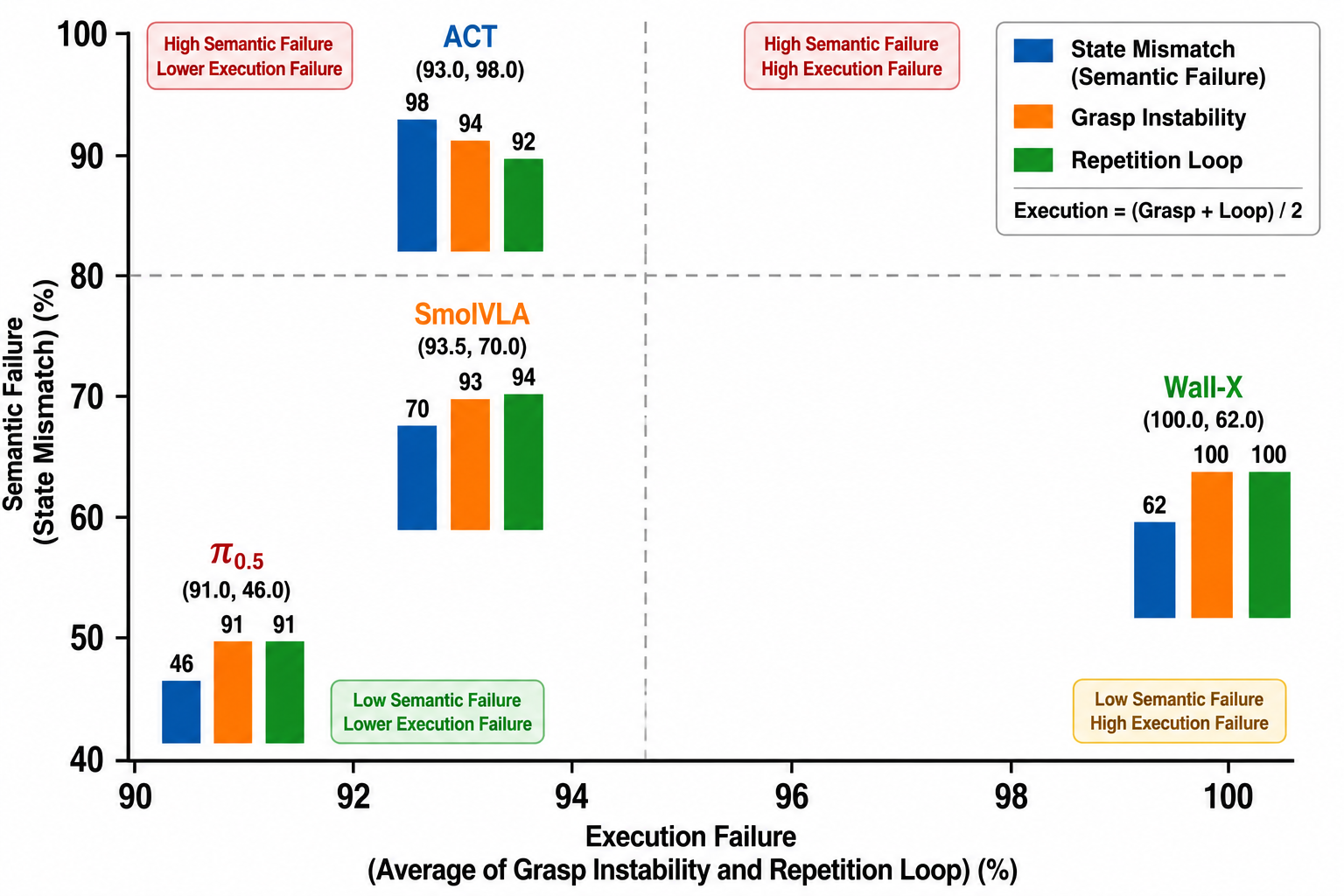}
    \caption{Semantic-level and execution-level failure analysis.}
    \label{fig:semantic}
\end{figure}

Figure~\ref{fig:semantic} illustrates the distribution of semantic-related and execution-related failures across all evaluated policies. Overall, execution-related failures remain consistently high across all methods, indicating that low-level manipulation robustness remains a major bottleneck for real-world deployment. Even policies with relatively strong semantic understanding continue to exhibit substantial execution instability.

In contrast, semantic-related failures show greater variation across models. ACT exhibits high failure rates in both categories, suggesting limitations in both scene understanding and action execution. $\pi_{0.5}$ achieves the lowest semantic failure rate while maintaining relatively high execution failure, indicating stronger semantic grounding but persistent challenges in control robustness. SmolVLA and Wall-X fall into intermediate regimes: SmolVLA shows higher semantic failure, whereas Wall-X exhibits lower semantic failure but remains dominated by execution-related failures.

\subsection{Recovery Behavior}

\begin{table}[t]
\centering
\caption{Failure recovery rate (\%) across evaluated models.}
\label{tab:recovery_rate}
\begin{tabular}{lc}
\toprule
Model & Recovery Rate (\%) \\
\midrule
 ACT & 6.45\% \\
SmolVLA     & 3.23\% \\
Wall-X      & 20.51\% \\
$\pi_{0.5}$ & 30.77\% \\
\bottomrule
\end{tabular}
\end{table}
Beyond task success, we evaluate the ability of policies to recover from intermediate execution failures without external intervention, which is critical for robust long-horizon manipulation in unstructured environments. Table~\ref{tab:recovery_rate} reports recovery rates computed over all identified recovery opportunities.

The results reveal clear differences in recovery capability across models. Among all evaluated methods, $\pi_{0.5}$ achieves the highest recovery rate (30.77\%), followed by Wall-X (20.51\%). In contrast, ACT (6.45\%) and SmolVLA (3.23\%) exhibit substantially lower recovery rates, indicating limited ability to self-correct once execution failures occur.
These results suggest that beyond initial action accuracy, recovery capability is an important factor differentiating manipulation policies in maintaining task progress under failure conditions.

\subsection{Episode-Level Summary of Task Performance}

\begin{table*}[t]
\centering
\caption{Episode-level summary of task performance, failure modes, and recovery behaviors across different models.}
\label{tab:detailed_analysis}

\resizebox{\textwidth}{!}{
\begin{tabular}{llccccccc}
\toprule
\textbf{Model} &
\textbf{Task} &
\textbf{Success} &
\textbf{Grasp} &
\textbf{Repetition} &
\textbf{State} &
\textbf{Precision} &
\textbf{Recovery} &
\textbf{Recovered} \\
&
&
\textbf{Episodes} &
\textbf{Instability} &
\textbf{Loop} &
\textbf{Mismatch} &
\textbf{Misalignment} &
\textbf{Opportunities} &
\textbf{Successes} \\
\midrule

\multirow{4}{*}{ACT}
& Pen Transfer              & 15 & 5 & 5  & 5  & 0 & 3  & 0 \\
& Color Sorting             & 0  & 19 & 19 & 20 & 6 & 7 & 0 \\
& Multi-object Packing      & 2  & 18 & 18 & 18 & 0 & 17 & 0 \\
& Precision Pen Placement   & 10 & 8  & 7  & 9  & 2 & 4  & 2 \\
\midrule

\multirow{4}{*}{SmolVLA}
& Pen Transfer              & 14 & 6 & 6 & 3 & 0 & 4 & 0 \\
& Color Sorting             & 1  & 18 & 18 & 14 & 1 & 8 & 1 \\
& Multi-object Packing      & 2  & 18 & 18 & 18 & 0 & 16 & 0 \\
& Precision Pen Placement   & 9  & 8 & 9 & 3 & 3 & 3 & 0 \\
\midrule

\multirow{4}{*}{Wall-X}
& Pen Transfer              & 19 & 1 & 1 & 1 & 0 & 1 & 1 \\
& Color Sorting             & 0  & 20 & 20 & 8 & 0 & 16 & 0 \\
& Multi-object Packing      & 6  & 14 & 14 & 14 & 0 & 17 & 5 \\
& Precision Pen Placement   & 16 & 4 & 4 & 1 & 0 & 5 & 2 \\
\midrule

\multirow{4}{*}{$\pi_{0.5}$}
& Pen Transfer              & 19 & 0 & 0 & 0 & 0 & 1 & 1 \\
& Color Sorting             & 2  & 18 & 17 & 5 & 1 & 8 & 0 \\
& Multi-object Packing      & 11 & 9 & 9 & 9 & 0 & 12 & 6 \\
& Precision Pen Placement   & 13 & 5 & 6 & 2 & 4 & 5 & 1 \\
\bottomrule
\end{tabular}
}
\end{table*}

Table~\ref{tab:detailed_analysis} provides a fine-grained episode-level decomposition of task outcomes, failure modes, and recovery behaviors across different policies and tasks. Compared to aggregate success metrics, it enables a more detailed understanding of \emph{where} and \emph{why} failures occur, as well as how different models behave after encountering execution errors.

Across all models and tasks, a consistent pattern emerges: execution-related failures (grasp instability, repetition loops, and state mismatches) occur substantially more frequently than precision misalignment, suggesting that instability in low-level control and state tracking remains a key challenge in long-horizon manipulation.

A second key observation is that failure modes exhibit a clear task-dependent structure. In simpler tasks such as \textit{Pen Transfer}, most models show relatively low failure counts and higher success rates, whereas more compositional tasks such as \textit{Multi-object Packing} and \textit{Color Sorting} induce higher frequencies of repetition loops and state mismatches, indicating that increased task complexity amplifies both reasoning and execution failures.

Recovery behavior further differentiates the evaluated policies. Although recovery opportunities are present across all tasks, only a subset are successfully converted into recovered successes. $\pi_{0.5}$ and Wall-X demonstrate comparatively stronger recovery effectiveness, particularly in higher-complexity tasks, suggesting better utilization of intermediate task states. In contrast, ACT and SmolVLA exhibit limited recovery success despite comparable or even higher numbers of recovery opportunities, indicating difficulty in exploiting partially successful states once failures occur.

Overall, the table highlights three key insights: (i) execution instability is the dominant failure mode across all models, (ii) task complexity is associated with increased failure frequency across both semantic and execution dimensions, and (iii) recovery capability serves as an important differentiating factor among policies beyond initial success rates.

\subsection{Success and Failure Examples}

We present representative success and failure cases for the multi-object packing task to provide qualitative insights into policy behavior beyond quantitative metrics. Success examples are selected from a strong-performing policy ($\pi_{0.5}$), while failure examples are drawn from a representative lower-performing baseline (ACT), highlighting contrasting execution patterns under identical task conditions. Task success and failure are defined according to the task-specific completion criteria in Table~\ref{tab:task_definition}. A trial is considered successful if all objects are correctly packed into the target box without violating task constraints; otherwise, it is considered a failure.

\begin{figure}[t]
\centering
\setlength{\tabcolsep}{2pt}

\begin{tabular}{cccc}

\multicolumn{4}{c}{\textbf{Multi-object Packing (Success)}} \\
\includegraphics[width=0.23\linewidth]{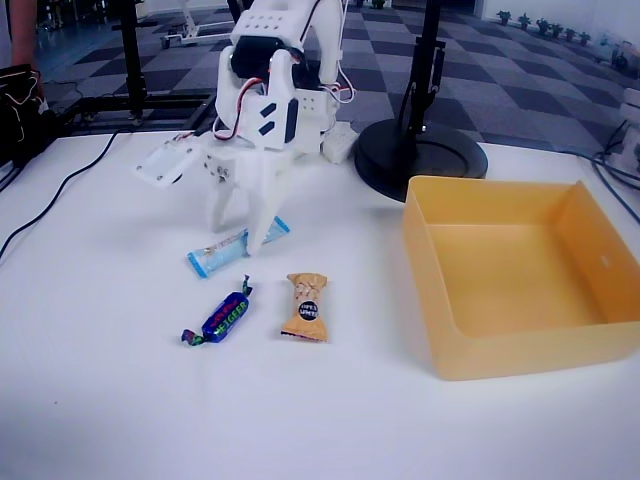}
& \includegraphics[width=0.23\linewidth]{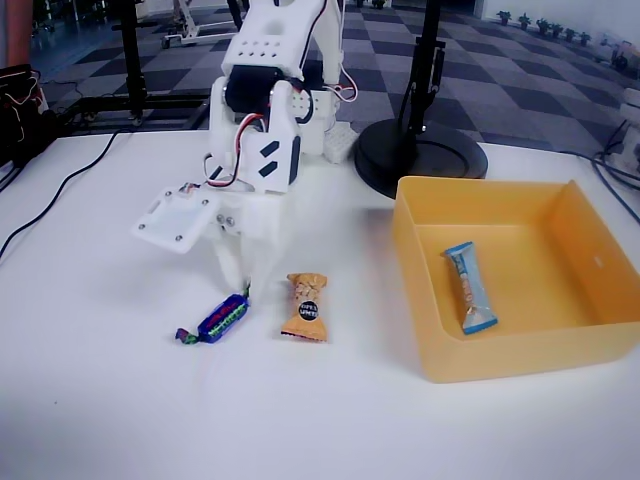}
& \includegraphics[width=0.23\linewidth]{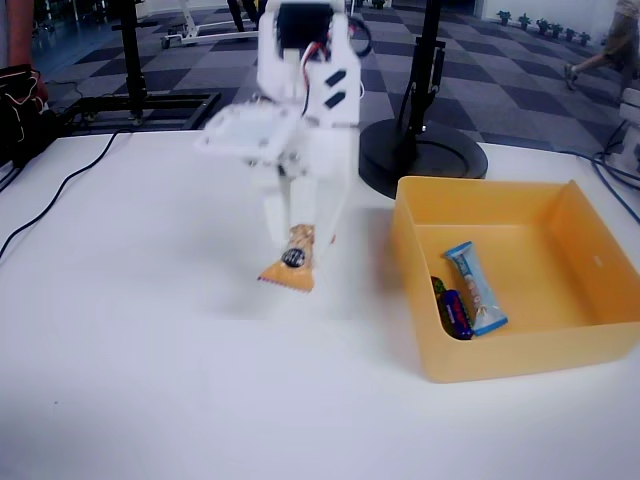}
& \includegraphics[width=0.23\linewidth]{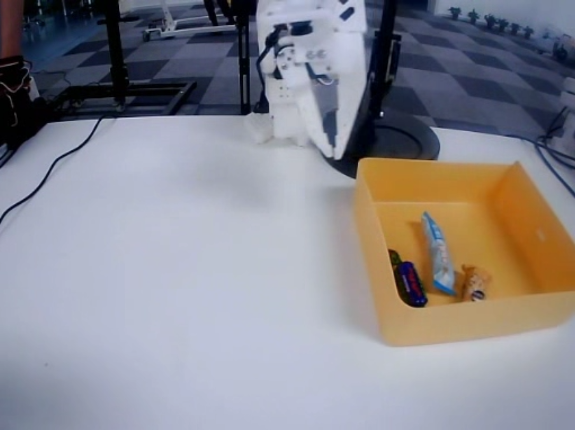}
\\

%\midrule

\multicolumn{4}{c}{\textbf{Multi-object Packing (Failure)}} \\
\includegraphics[width=0.23\linewidth]{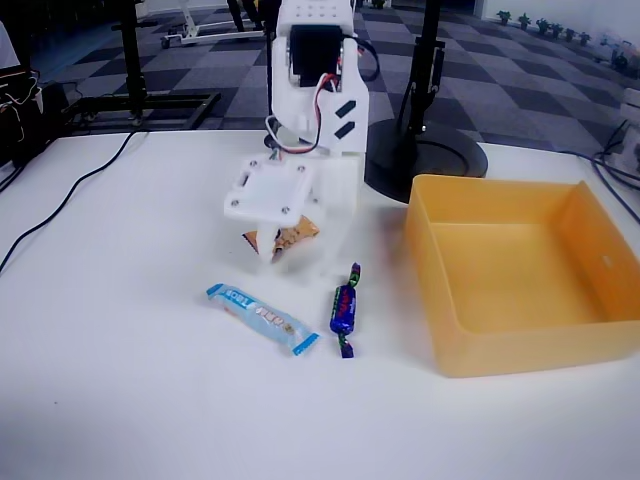}
& \includegraphics[width=0.23\linewidth]{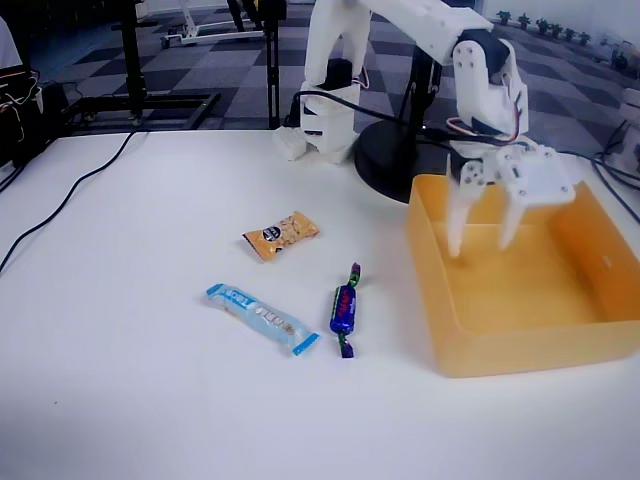}
& \includegraphics[width=0.23\linewidth]{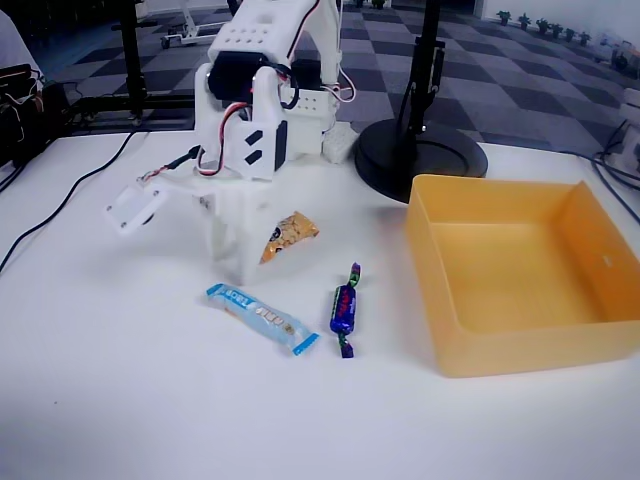}
& \includegraphics[width=0.23\linewidth]{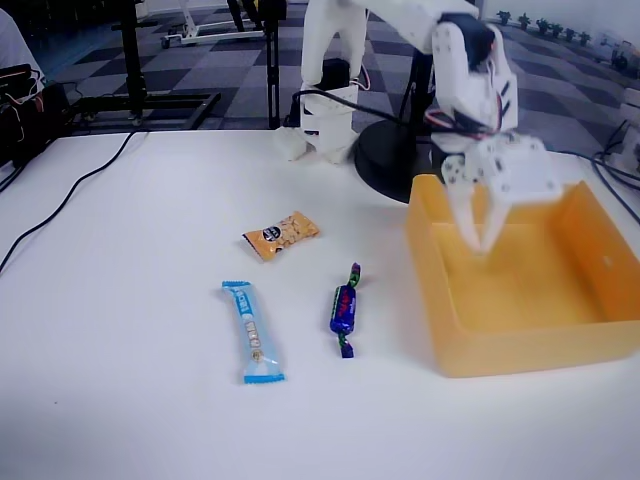}
\\

\end{tabular}

\caption{
Example trajectories for the Multi-object Packing task.
Success cases are taken from the best-performing model, while failure cases are taken from a lower-performing model.
}
\label{fig:multi_object_packing}
\end{figure}

As shown in Figure~\ref{fig:multi_object_packing}, successful trajectories typically exhibit stable grasping, consistent object placement order, and effective recovery from intermediate misalignments, resulting in complete task execution. In contrast, failure cases often involve repeated grasp attempts, unstable object placement, and early-stage state misinterpretation, which lead to compounding errors and eventual task termination. These qualitative results are consistent with our quantitative findings in Section~\ref{sec:failure_analysis}, where execution instability and state mismatch are identified as dominant failure modes, while recovery capability differentiates stronger policies from weaker baselines.

\section{Discussion}

In this paper, we present a standardized benchmark for evaluating Vision-Language-Action (VLA) models on the low-cost SO-101 robotic platform. Beyond conventional task success rates, the benchmark incorporates a structured failure taxonomy, semantic- and execution-level failure decomposition, and recovery-aware evaluation metrics to provide a more comprehensive characterization of embodied policy robustness.

\subsection{Low-Cost Robotics as a Stress Test for VLA Models}

Our experimental results suggest that low-cost robotic platforms such as SO-101 provide a valuable stress-test environment for Vision-Language-Action (VLA) models, exposing failure behaviors that are often hidden in controlled or low-noise evaluation settings. Embodiment uncertainty associated with low-cost hardware poses additional challenges for control stability, language grounding, and long-horizon execution consistency, revealing robustness limitations that may not be fully captured by conventional simulation benchmarks. This observation is further supported by our failure mode analysis and recovery evaluation, which reveal consistent differences in policy behavior under identical real-world deployment conditions. Consequently, low-cost robotic platforms provide complementary evaluation signals beyond aggregate task success, particularly with respect to execution stability, failure recovery, and robustness under hardware-induced stochasticity.

\subsection{Structured Failure Patterns Across Policies}

A key finding of this study is that policy failures exhibit systematic patterns rather than random execution errors, with distinct failure characteristics emerging across different policy architectures under identical deployment conditions. Execution-related failures, particularly repetition loops and grasp instability, remain common across all evaluated policies, while recovery capability varies substantially across architectures. In contrast, the imitation learning baseline exhibits comparatively higher state mismatch rates, indicating weaker consistency between its internal task representation and the observed environment state. These observations suggest that different policy architectures are associated with distinct failure behaviors under real-world embodiment constraints. More importantly, failure mode analysis provides complementary information beyond aggregate success rates, as models with similar task completion performance may exhibit substantially different failure dynamics and recovery characteristics.

\subsection{Gap Between Success Rate and Embodied Robustness}

High task success rates alone are insufficient to characterize embodied policy robustness. Our experimental results reveal a noticeable gap between task completion performance and robustness-related indicators such as failure mode composition and recovery capability. In particular, models with comparable success rates can exhibit substantially different recovery behaviors and underlying failure characteristics, indicating that binary success metrics alone provide only a limited view of policy robustness.
Moreover, the proposed failure analysis reveals systematic differences in execution stability and semantic consistency across evaluated policies, suggesting that similar overall success rates may correspond to different execution patterns during task completion. These observations highlight the importance of incorporating complementary evaluation metrics, including structured failure analysis and recovery behavior, into future embodied AI benchmarks.
Overall, our results suggest that evaluating failure dynamics and recovery capability, in addition to conventional success rates, provides a more comprehensive assessment of embodied policy robustness under real-world low-cost robotic deployment conditions.

\section{Limitations and Future Work}

Despite the encouraging findings, this study has several limitations. First, all experiments are conducted on a single low-cost robotic platform (SO-101), and the observed failure characteristics may not fully generalize to robots with different embodiment properties, sensing modalities, or control interfaces. Second, the proposed benchmark focuses on tabletop manipulation tasks and does not cover more complex embodied scenarios such as mobile manipulation, bimanual coordination, or long-horizon multi-stage activities. Third, although representative state-of-the-art Vision-Language-Action (VLA) policies are evaluated, the rapidly evolving nature of embodied foundation models suggests that additional architectures should be incorporated in future studies. 

\subsection{Benchmark Scaling and Task Diversity}

An important direction for future work is to extend the proposed benchmark to a broader spectrum of embodied manipulation scenarios. Incorporating deformable objects, contact-rich manipulation, cluttered environments, articulated objects, and multi-stage assembly tasks would provide a more comprehensive evaluation of policy generalization under diverse sources of embodiment uncertainty. Expanding both task complexity and environmental diversity would also facilitate more systematic analysis of long-horizon failure accumulation and recovery behavior.

\subsection{Toward Recovery-Aware Embodied Intelligence}

Our experimental results show that execution instability remains the dominant source of policy failure, while recovery capability varies substantially across different architectures and contributes significantly to overall task robustness. These observations suggest that recovery should be treated as a first-class capability in future Vision-Language-Action systems rather than as an implicit byproduct of policy execution. Promising research directions include recovery-aware policy learning, failure-conditioned planning, corrective feedback mechanisms, adaptive replanning, and training objectives that explicitly encourage robust long-horizon execution under embodiment uncertainty.

\subsection{Cross-Platform Evaluation and Robustness Assessment}

Future work should further investigate policy behavior across multiple robotic platforms with different embodiment characteristics, sensing configurations, and control precision. Such cross-platform evaluation would help distinguish hardware-specific artifacts from more fundamental limitations of current policy architectures, while providing a more comprehensive assessment of robustness and generalization in real-world embodied AI systems. More broadly, integrating failure mode analysis and recovery evaluation into future benchmark protocols may provide complementary evaluation signals beyond conventional task success rates, leading to more informative assessments of embodied intelligence.

\bibliographystyle{ACM-Reference-Format}
\bibliography{sample-base}

@article{palm_e,
  title={PaLM-E: An Embodied Multimodal Language Model},
  author={Driess, Danny and Xia, Fei and Sajjadi, Mehdi S M and others},
  journal={arXiv preprint arXiv:2303.03378},
  year={2023}
}

@inproceedings{rt2,
  title={RT-2: Vision-Language-Action Models Transfer Web Knowledge to Robotic Control},
  author={Zitkovich, Blake and Yu, Tianli and Xu, Sherry and others},
  booktitle={Conference on Robot Learning (CoRL)},
  year={2023},
  pages={2165--2183}
}

@article{octo,
  title={Octo: An Open-Source Generalist Robot Policy},
  author={Octo Model Team and Ghosh, D and Walke, H and others},
  journal={arXiv preprint arXiv:2405.12213},
  year={2024}
}

@article{openvla,
  title={OpenVLA: An Open-Source Vision-Language-Action Model},
  author={Kim, M. J. and Pertsch, K. and Karamcheti, S. and others},
  journal={arXiv preprint arXiv:2406.09246},
  year={2024}
}

@article{openxembodiment,
  title={Open X-Embodiment: Robotic Learning Datasets and RT-X Models},
  author={{Open X-Embodiment Collaboration} and others},
  journal={arXiv preprint arXiv:2310.08864},
  year={2023}
}

@article{rlbench,
  title={RLBench: The Robot Learning Benchmark \& Learning Environment},
  author={James, Stephen and Ma, Zicong and Arrojo, David Rovick and Davison, Andrew J},
  journal={IEEE Robotics and Automation Letters},
  volume={5},
  number={2},
  pages={3019--3026},
  year={2020},
  doi={10.1109/LRA.2020.2974707}
}

@inproceedings{robosuite,
  title={robosuite: A Modular Simulation Framework and Benchmark for Robot Learning},
  author={Zhu, Yuke and Wong, Josiah and Mandlekar, Ajay and others},
  booktitle={Conference on Robot Learning (CoRL)},
  year={2020},
  howpublished={\url{https://arxiv.org/abs/2009.12293}}
}

@article{act,
  title={Learning Fine-Grained Bimanual Manipulation with Low-Cost Hardware},
  author={Zhao, Tony Z and Kumar, Vikash and Levine, Sergey and others},
  journal={arXiv preprint arXiv:2304.13705},
  year={2023}
}

@article{pi05,
  title={$\\pi_{0.5}$: A Vision-Language-Action Model with Open-World Generalization},
  author={Physical Intelligence and Black, Kevin and Brown, Noah and others},
  journal={arXiv preprint arXiv:2504.16054},
  year={2025}
}

@article{smolvla,
  title={SmolVLA: A Vision-Language-Action Model for Affordable and Efficient Robotics},
  author={Shukor, Mustafa and Aubakirova, Dana and Capuano, Francesco and Kooijmans, Pepijn and Palma, Steven and Zouitine, Adil and Aractingi, Michel and Pascal, Caroline and Russi, Martino and Marafioti, Andres and others},
  journal={arXiv preprint arXiv:2506.01844},
  year={2025}
}

@misc{wallx,
  title={Building General-Purpose Robots Based on Embodied Foundation Models},
  author={{X-Square Robot Team}},
  howpublished={\url{https://github.com/X-Square-Robot/wall-x}},
  year={2025}
}

@misc{so101,
  title={SO-101 Low-Cost Robotic Manipulation Platform},
  author={{LeRobot Community}},
  howpublished={\url{https://github.com/huggingface/lerobot}},
  year={2024},
  note={Open-source low-cost robot platform used for embodied AI research}
}

@misc{franka_panda,
  title={Franka Emika Panda Robot},
  author={{Franka Emika GmbH}},
  year={2020},
  howpublished={\url{https://www.franka.de/technology}},
  note={Accessed: 2026-05-28}
}

@misc{widowx,
  title={WidowX Robot Arm},
  author={{Trossen Robotics}},
  year={2020},
  howpublished={\url{https://www.trossenrobotics.com/widowxrobotarm}},
  note={Accessed: 2026-05-28}
}

@misc{so101_dataset,
  title={SO-101 400-Demonstrations VLA Evaluation Dataset},
  author={Qiu, Xinchuan and Yu, Yi},
  year={2026},
  howpublished={\url{https://huggingface.co/collections/Qiu-Xinchuan/400-so-101-vla-evaluate-dataset}}
}

% \appendix

\end{document}